\documentclass[10pt,twocolumn,letterpaper]{article}
\usepackage{cvpr}              
\usepackage{graphicx}

\makeatletter
\@namedef{ver@everyshi.sty}{}
\makeatother
\usepackage{tikz}

\usepackage{comment}
\usepackage{amsmath,amssymb} 
\usepackage{color}

\usepackage{adjustbox}
\usepackage{epsfig}
\usepackage{amsmath}
\usepackage{amssymb}
\usepackage{mathtools}
\usepackage{booktabs}

\usepackage[ruled, noend, noline]{algorithm2e}
\usepackage{algorithmic}
\usepackage{multirow}
\usepackage{threeparttable}

\usepackage[pagebackref,breaklinks,colorlinks]{hyperref}

\usepackage[capitalize]{cleveref}
\crefname{section}{Sec.}{Secs.}
\Crefname{section}{Section}{Sections}
\Crefname{table}{Table}{Tables}
\crefname{table}{Tab.}{Tabs.}

\def\cvprPaperID{*****} 
\def\confName{CVPR}
\def\confYear{2022}

\begin{document}

\title{AutoAlignV2: Deformable Feature Aggregation for Dynamic \\Multi-Modal 3D Object Detection}

\author{Zehui Chen$^1$ \quad Zhenyu Li$^{2}$ \quad Shiquan Zhang$^{3}$ \quad Liangji Fang$^{3}$ \quad Qinhong Jiang$^{3}$ \quad Feng Zhao$^1$ \\
$^{1}$University of Science and Technology of China \\ $^{2}$Harbin Institute of Technology  \quad $^{3}$SenseTime Research}

\maketitle

\begin{abstract}
	Point clouds and RGB images are two general perceptional sources in autonomous driving. The former can provide accurate localization of objects, and the latter is denser and richer in semantic information. Recently, AutoAlign \cite{chen2022autoalign} presents a learnable paradigm in combining these two modalities for 3D object detection. However, it suffers from high computational cost introduced by the global-wise attention. To solve the problem, we propose Cross-Domain DeformCAFA module in this work. It attends to sparse learnable sampling points for cross-modal relational modeling, which enhances the tolerance to calibration error and greatly speeds up the feature aggregation across different modalities. To overcome the complex GT-AUG under multi-modal settings, we design a simple yet effective cross-modal augmentation strategy on convex combination of image patches given their depth information. Moreover, by carrying out a novel image-level dropout training scheme, our model is able to infer in a dynamic manner. To this end, we propose AutoAlignV2, a faster and stronger multi-modal 3D detection framework, built on top of AutoAlign. Extensive experiments on nuScenes benchmark demonstrate the effectiveness and efficiency of AutoAlignV2. Notably, our best model reaches 72.4 NDS on nuScenes test leaderboard, achieving new state-of-the-art results among all published multi-modal 3D object detectors. Code will be available at \href{https://github.com/zehuichen123/AutoAlignV2}{https://github.com/zehuichen123/AutoAlignV2}.
\end{abstract}


\section{Introduction}

\begin{figure}[ht!]
	\center
   \includegraphics[width=0.5\textwidth]{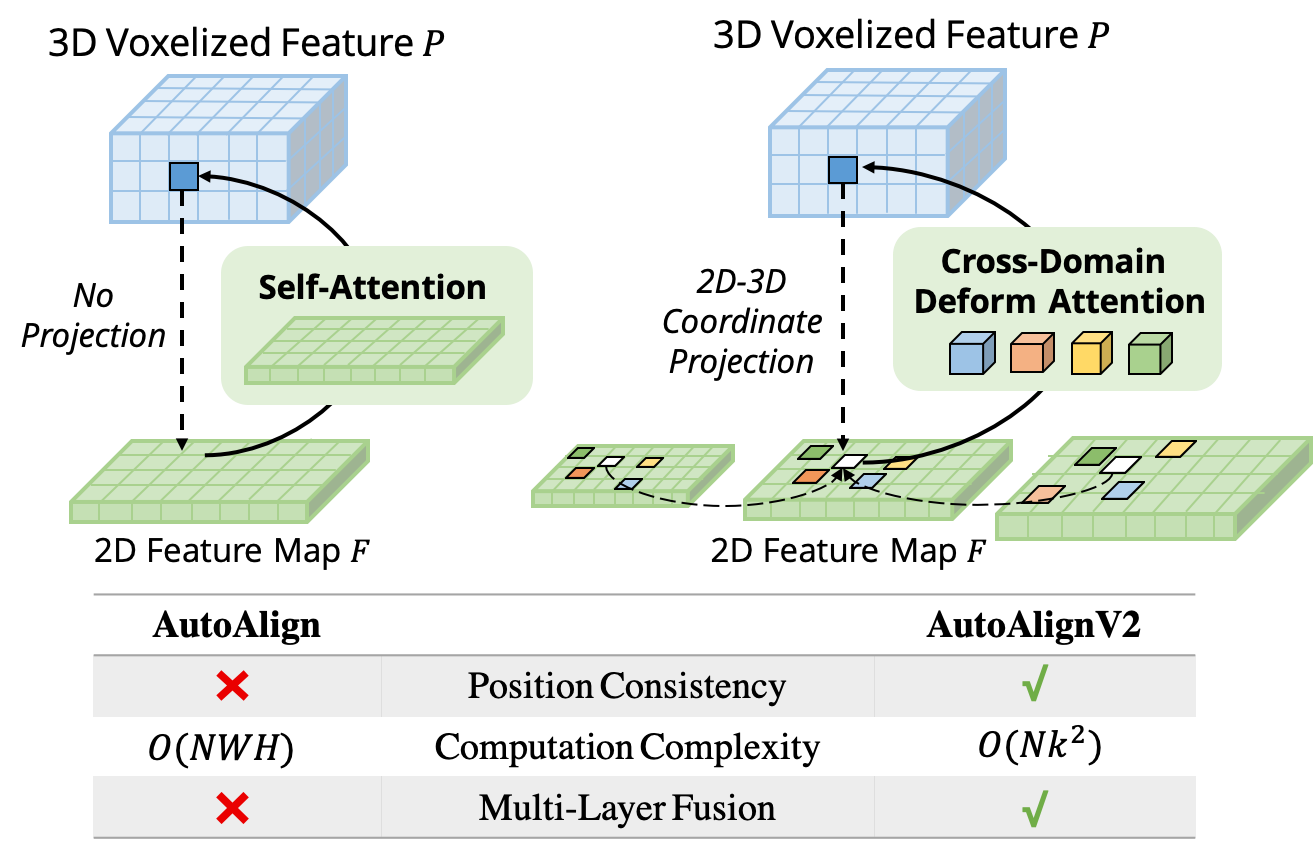}
\caption{The comparison between AutoAlignV2 and AutoAlign. AutoAlignV2 hints at the alignment module with general mapping relationship guaranteed by deterministic projection matrix, and simultaneously reserves the ability to automatically adjust the positions of feature aggregation. Due to the lightweight computational cost, AutoAlignV2 is able to aggregate multi-layer features for hierarchical imagery information. 
}
\vskip -1em
\label{fig:compare}
\end{figure}

3D object detection serves as a fundamental computer vision task in autonomous driving. Modern 3D object detectors~\cite{yan2018second,lang2019pointpillars,shi2020pv,mao2021voxel} have demonstrated promising performance on competitive benchmarks including KITTI \cite{geiger2012we}, Waymo~\cite{sun2020scalability}, and nuScenes~\cite{caesar2020nuscenes} datasets. Despite the rapid progress in detection accuracy, the room for further improvement is still large. Recently, an upsurging stream in combining RGB images and LiDAR points for accurate detection has drawn many attentions~\cite{wang2021pointaugmenting,zhang2020exploring,li2022simipu,li2022deepfusion,bai2022transfusion,liu2022bevfusion}. Different from the point clouds which are beneficial for spatial localization, imagery data are more superior in providing semantic and textural information, \textit{i.e.}, more suitable for classification. Therefore, it is believed that these two modalities are complementary to each other and can further promote the detection accuracy. 

However, how to effectively combine these heterogeneous representations for 3D object detection has not been fully explored. In this work, we mainly attribute the current difficulties of training cross-modal detectors to two aspects.
On one hand, the fusion strategy in combining imagery and spatial information remains sub-optimal. Due to the heterogeneous representations between RGB images and point clouds, features need to be carefully aligned before being aggregated together. This is often achieved by establishing deterministic correspondence between the point and the image pixel through LiDAR-camera projection matrix~\cite{sindagi2019mvx,wang2021pointaugmenting,zhang2020exploring}. AutoAlign \cite{chen2022autoalign} proposes a learnable global-wise alignment module for automatic registration and achieves good performance. However, it has to be trained with the help of CSFI module to acquire the inner positional matching relationship between points and image pixels. Besides, the complexity of attention-style operation is quadratic to the image size, making it impractical to apply queries on high-resolution feature maps (\textit{e.g.,} $C_2, C_3$). Such a limitation can lead to coarse and inaccurate image information and the loss of hierarchical representations brought by FPN (See Figure \ref{fig:compare}). 
On the other hand, data augmentation, especially GT-AUG \cite{yan2018second}, is a crucial step for 3D detectors to achieve competitive results. In terms of multi-modal methods, an important problem is how to keep synchronization between images and point clouds when conducting cut and paste operations. MoCa \cite{zhang2020exploring} uses labor-intensive mask annotations in 2D domain for accurate image features. Box-level annotations are also applicable but delicate and complex points filtering is required \cite{wang2021pointaugmenting}. 

\begin{figure*}[t!]
   \includegraphics[width=1.0\textwidth]{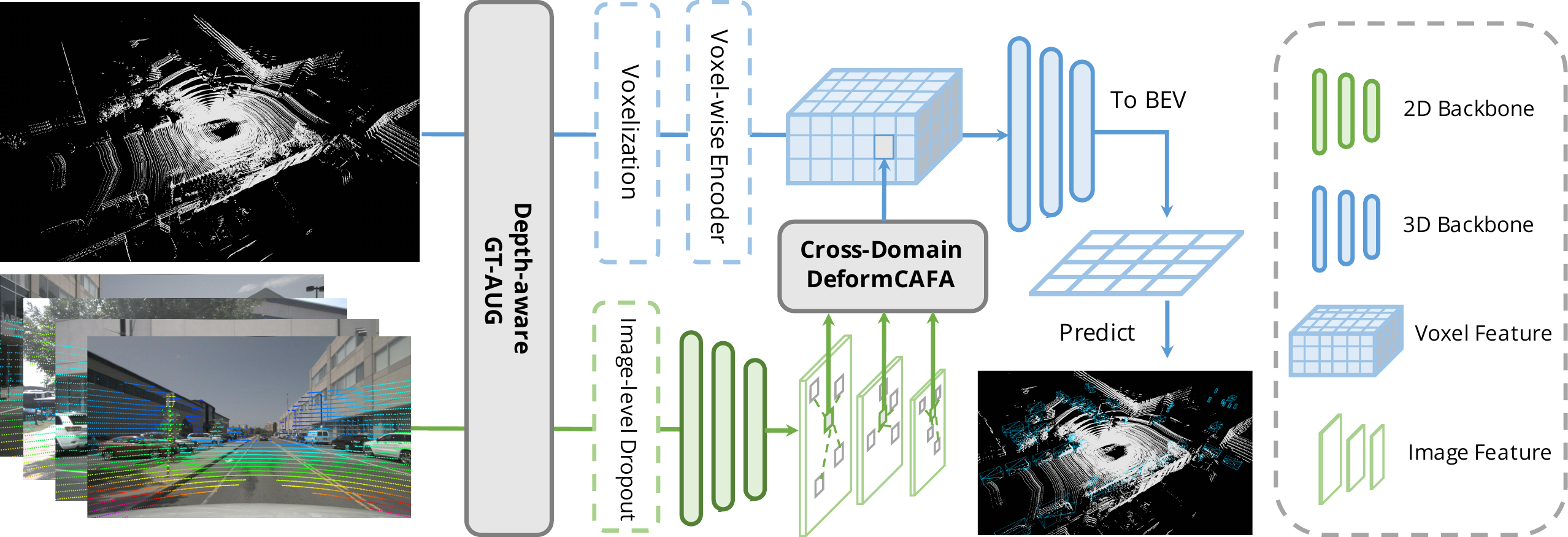}
\caption{The overall framework of AutoAlignV2. It differs from AutoAlign in three aspects: (i) the proposed Cross-Domain DeformCAFA module enhances the representations with better imagery features and improves the efficiency of the fusion process, (ii) the Depth-Aware GT-AUG algorithm greatly simplifies the synchronization issue among 2D-3D joint augmentations, and (iii) the adoption of image-level dropout training strategy enables our model to infer in a dynamic fusion manner.
}
\label{fig:intro_pdf}
\end{figure*}

In this work, we propose AutoAlignV2 to mitigate the aforementioned issues in a much simpler and more effective way. It hints at the alignment module with the general mapping relationship guaranteed by deterministic projection matrix and simultaneously reserves the ability to automatically adjust the positions of feature aggregation. As for the synchronization issue in 2D-3D joint augmentation, a novel depth-aware GT-AUG algorithm is introduced to cope with object occlusion in the image domain, getting rid of the complex point cloud filtering or the need for delicate mask annotations. We also present a new training scheme named image-level dropout strategy, which enables the model to infer results dynamically even without images. Through extensive experiments, we validate the effectiveness of AutoAlignV2 on two representative 3D detectors: Object DGCNN \cite{wang2021object} and CenterPoint \cite{yin2021center}, and achieve new state-of-the-art performance on the competitive nuScenes benchmark.

\section{Related Work}
\subsection{Object Detection with Point Cloud}
Existing 3D object detectors can be broadly categorized as point-based and voxel-based approaches. Point-based methods directly predict the regression boxes from points~\cite{yang2019std,shi2020points}. For example, Point R-CNN~\cite{shi2019pointrcnn} adopts a semantic network to segment the point clouds and then generates the proposals at each foreground point. 3DSSD~\cite{yang20203dssd} fully applies point-level predictions on the one-stage architecture, where an anchor-free head is designed after the PointNet-like feature extraction. Although the accurate 3D localization information is maintained, these algorithms often suffer from high computational cost~\cite{shi2020pv}. Different from the point-wise detection, voxel-based approaches transform sets of unordered points into 2D feature map through voxelization, which can be directly applied with convolutional neural networks~\cite{zhou2018voxelnet,qi2019deep,deng2021voxel}. For instance, VoxelNet~\cite{zhou2018voxelnet} is a widely-used paradigm where a VFE layer is proposed to extract unified features for each 3D voxel. Based on this, CenterPoint~\cite{yin2021center} presents a center-based label assignment strategy, achieving competitive performance in 3D object detection. 

\subsection{Multi-Modal 3D Object Detection}
Recently, there has been an increasing attention on multi-modal data for 3D object detection~\cite{liang2018deep,pang2020clocs}. AVOD~\cite{ku2018joint} and MV3D~\cite{chen2017multi} are two pioneer works in this field, where 2D and 3D RoI are directly concatenated before box prediction. Qi \textit{et al.}~\cite{qi2018frustum} utilized images to generate 2D proposals and then lifted them up to 3D space (frustum), which narrows the searching space in point clouds. 3D-CVF~\cite{yoo20203d} and EPNet~\cite{huang2020epnet} explore the fusion strategy on feature maps across different modalities with a learned calibration matrix. Though easy-to-implement, they are likely to suffer from coarse feature aggregation. To mitigate this issue, various approaches~\cite{sindagi2019mvx,zhang2020exploring,vora2020pointpainting} fetch pixel-wise image features with camera-LiDAR projection matrix given by 3D coordinates. As an example, MVX-Net~\cite{sindagi2019mvx} provides an easy-to-extend framework for cross-modal 3D object detection with joint optimization in 2D and 3D branches. AutoAlign~\cite{chen2022autoalign} formulates the projection relationship as an attention map and automates the learning of such an alignment through the network. In this work, we explore a faster and more efficient alignment strategy to further boost the performance of point-wise feature aggregation.
\section{AutoAlignV2}
The aim of AutoAlignV2 is to effectively aggregate image features for further performance enhancement of 3D object detectors. We start with the basic architecture of AutoAlign: the paired images are input into a light-weight backbone, ResNet~\cite{wang2020cspnet}, followed by FPN~\cite{liu2018path} to get the feature maps. Then, relevant imagery information is aggregated through a learnable alignment map to enrich the 3D representations of non-empty voxels during the voxelization phase. Finally, the enhanced features will be fed into the subsequent 3D detection pipeline to generate the instance predictions. 

Such a paradigm could aggregate heterogeneous features in a data-driven way. However, there are two main bottlenecks that still hinder the performance. The first one is inefficient feature aggregation. Although global-wise attention map automates the feature alignment between RGB images and LiDAR points, the computational cost is high: given the voxel number $N$ and the size of image feature $W\times H$, the complexity is $O(NWH)$. Due to the large value of $WH$, AutoAlign discards other layers except $C_5$ to reduce the cost. The second one is complex data augmentation synchronization between image and points. GT-AUG is an essential step for high-performance 3D object detectors, but how to keep the semantic consistency between the points and the image during training remains a complicated problem. 

In this section, we show that the aforementioned challenges can be effectively resolved through the proposed AutoAlignV2, which consists of two parts: Cross-Domain DeformCAFA module and Depth-Aware GT-AUG data augmentation strategy (see Figure \ref{fig:intro_pdf}). We also present a novel image-level dropout training strategy, which enables our model to infer in a more dynamic manner. 

\subsection{Deformable Feature Aggregation}

\subsubsection{Revisiting to CAFA}
We first revisit the Cross-Attention Feature Alignment module proposed in AutoAlign. Instead of establishing deterministic correspondence with the camera-LiDAR projection matrix, it models the mapping relationship with a learnable alignment map, which enables the network to automate the alignment of non-homogenous features in a dynamic and data-driven manner. Specifically, given the feature map $F = \{f_1, f_2,...,f_{hw}\}$($f_i$ indicates the image feature of the $i^{th}$ spatial position) and voxel features $P = \{p_1, p_2, ..., p_J\}$ ($p_j$ indicates each non-empty voxel feature) extracted from raw point clouds, each voxel feature $p_j$ will query the whole image pixels and generate the attention weights based on the dot-product similarity between the voxel feature and the pixel feature. The final output of each voxel feature is the linear combination of values on all the pixel features according to the attention weights. Such a paradigm enables the model to aggregate semantically relevant spatial pixels to update $p_j$ and demonstrates superior performance compared to bilinear interpolation of features. However, the huge computational cost limits the query candidate to $C_5$ only, losing the fine-grained information from high-resolution feature maps. 

\subsubsection{Cross-Domain DeformCAFA}
\begin{figure}[!h]
    \centering
    \includegraphics[width=0.45\textwidth]{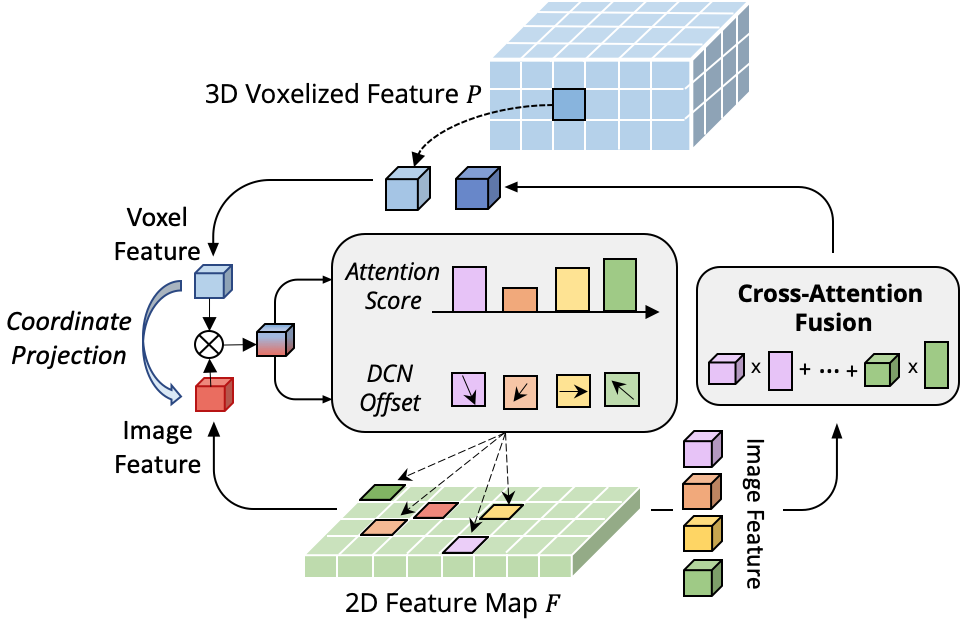}
    \caption{Illustration of the Cross-Domain DeformCAFA module. It first combines coordinate-corresponding voxel and image features to generate cross-domain tokens, which are then used to guide the aggregated positions in 2D feature map through learnable convolutional offset. The final fused feature is obtained by the cross-attention fusion of aggregated image feature and original voxel feature.}
    \label{fig:deformcafa}
\end{figure}

The bottleneck of CAFA is that it takes all the pixels as possible spatial positions. Based on the attributes of 2D images, the most relative information is mainly located at \textit{geometrically-nearby} locations. Therefore, it is unnecessary to consider all the positions but only several key-point regions. Inspired by this, we introduce a novel \textit{Cross-Domain DeformCAFA} operation (see Figure \ref{fig:deformcafa}), which greatly reduces the sampling candidates and dynamically decides the key-point regions on the image plane for each voxel query feature. 

More formally, given the feature map $\mathbf{F} \in \mathbb{R}^{h \times w \times d}$ extracted from the image backbone (\textit{e.g.,} ResNet, CSPNet) and non-empty voxel features $\mathbf{P} \in \mathbb{R}^{N\times c}$, we first compute the reference points $R_i = (r_x^x, r_y^i)$ in the image plane from each voxel feature center $V_i =(v_x^i, v_y^i, v_z^i)$ with the camera projection matrix $T_{cam-lidar}$,
\begin{equation}
	R_i = \mathbf{RC}\cdot T_{cam-lidar} \cdot V_i,
\end{equation} 
where $\mathbf{RC}$ is the combination of the rectifying rotation matrix and calibration matrix of the camera. After obtaining the reference point $R_i$, we adopt bilinear interpolation to get the feature $F_{i}$ in the image domain. The query feature $Q_i$ is derived as the element-wise product of the image feature $F_{i}$ and the corresponding voxel feature $P_j$ (to be discussed later). The final deformable cross-attention feature aggregation is calculated by,
\begin{equation}
\begin{split}
	\text{DeformCAFA}(Q_i, R_i, \mathbf{F}) = \hspace{110pt} \\
	\sum_{m=1}^{M} \mathbf{W}_m \left[\sum_{k=1}^K A_{mqk}(Q_i)\cdot \mathbf{W'}_m\mathbf{F}(R_i + \Delta R_{mqk})\right],
\end{split}
\end{equation}
where $\mathbf{W}_m$ and $\mathbf{W}'_m$ are learnable weights, and $A_{mqk}$ is a MLP to generate attention scores on the aggregated image features. Following the design of self-attention mechanism, we adopt $M$ attention split heads. Here, $K$ is the number of sampling positions ($K^2 \ll HW$, \textit{e.g.,} $K=4$). With the help of dynamically generated sampling offset $\Delta_{mqk}$, DeformCAFA is able to conduct cross-domain relational modeling much faster than vanilla operation. The complexity is reduced from $O(NWH)$ to $O(NK^2)$, enabling us to perform multi-layer feature aggregation, \textit{i.e.,} to fully utilize the hierarchical information provided by FPN layers. Another advantage of DeformCAFA is that it explicitly maintains the positional consistency with the camera projection matrix to obtain the reference points. Hence, even without adopting the CFSI module proposed in AutoAlign, our DeformCAFA can yield a semantically and positionally consistent alignment.

\begin{figure}[!h]
    \centering
    \includegraphics[width=0.5\textwidth]{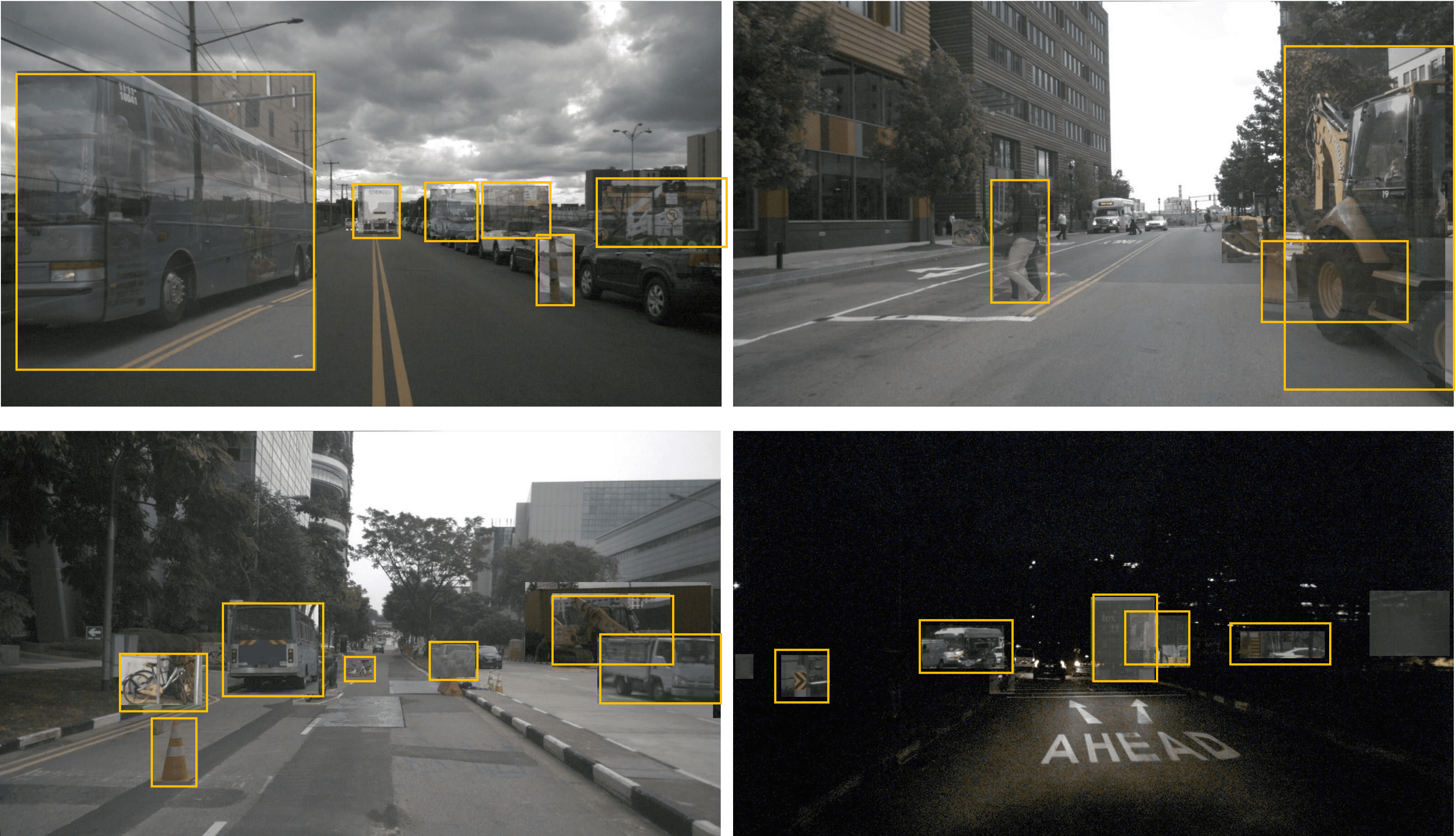}
    \caption{Visualization of the augmented images with the proposed Depth-Aware GT-AUG. The samples are randomly selected from nuScenes dataset.}
    \label{fig:aug_img}
\end{figure}

\subsubsection{Cross-Domain Token Generation}
\label{sec:cross_token}

The sparse-style DeformCAFA greatly improves the efficiency compared to vanilla non-local operation. However, when directly applying voxel features as the token to generate attention weights and deformable offsets, the detection performance is barely comparable to or even worse than its bilinear-interpolation counterparts. After careful analysis, we find a cross-domain knowledge translation issue in the token generation process. Different from the original deformable operation, which is usually performed under the unimodal setting~\cite{carion2020end,zhu2020deformable}, cross-domain attention requires information from both modalities. However, the voxel features that only consist of spatial representations, can hardly perceive information in the image domain. Therefore, allowing interaction between different modalities is of great importance. 

Motivated by \cite{li2021fully}, we hypothesize that the representation of each object can be explicitly disentangled into two components: the domain-specific and the instance-specific information. The former refers to the data related to the representation itself, including the built-in attributes of domain features, while the latter represents the identity information about the object, regardless of the domain it is encoded in. Concretely, given the corresponding paired image feature $F_i$ and voxel feature $P_j$, we have,
\begin{equation}
F_i = D^{2D}_i \cdot M_{obj}^i,~
P_j = D^{3D}_j \cdot M_{obj}^j, 
\end{equation}
where $D^{2D}_i$ and $D^{3D}_j$ are domain-related features in the image and point domains, while $M_{obj}^i$ and $M_{obj}^j$ are the object-specific representations, respectively. Since $F_i$ and $P_j$ are the geometrically-paired features, $M_{obj}^i$ and $M_{obj}^j$ can be close in the instance-specific representation space (\textit{i.e.,} $M_{obj} \approx M_{obj}^i \approx M_{obj}^j$). Based on this, we can implicitly interact features of different domain knowledges with,
\begin{equation}
    Token = f(F_i \cdot P_j) = f(D^{2D}_i\cdot D^{3D}_j \cdot (M_{obj})^2), 
\end{equation}
where $f$ is one fully connected (FC) layer to aggregate cross-domain information and improve the flexibility of token generation. 

\subsection{Depth-Aware GT-AUG}
\label{sec:depth_aug}

Data augmentation is a crucial part of achieving competitive results for most deep learning models. However, in terms of multi-modal 3D object detection, it is hard to keep synchronization between point clouds and images when combining them together in data augmentation, mainly due to object occlusions or changes in the viewpoints. To solve the problem, we design a simple yet effective cross-modal data augmentation named \textit{Depth-Aware GT-AUG}. Different from the methods described in \cite{wang2021pointaugmenting,zhang2020exploring}, our approach abandons the complex point cloud filtering process or the requirement of delicate mask annotation in the image domain. Instead, inspired by the MixUp proposed in \cite{zhang2017mixup}, we incorporate the depth information from 3D object annotations to mix up the image regions. 

Specifically, given the virtual objects $P$ to paste, we follow the same 3D implementation in GT-AUG \cite{yan2018second}. As for the image domain, we first sort them in a far-to-near order. For each to-paste object, we crop the same region from the original image and combine them with a mix-up ratio of $\alpha$ on the target image. The detailed implementation is shown in Algorithm \ref{algo:gt_aug}.  

\begin{algorithm} 
\caption{Depth-Aware GT-AUG}
\begin{algorithmic}[1]
\REQUIRE Object Points Set $\mathbf{P^{3D}}$, Object Image Patches Set $\mathbf{P^{2D}}$, Object Depths Set $\mathbf{D}$, Points $\mathbf{P}$, Image $\mathbf{I}.$
\STATE ObjectInds $\leftarrow$ AscendingSort($\mathbf{D}$);
\FORALL{$i$ such that $i$ $\in$ ObjectInds}
    \STATE $//$ point augmentation
    \STATE $\mathbf{P} \leftarrow \mathbf{P}$ + $\mathbf{P^{3D}_{i}}$;
    \STATE $//$ image augmentation
    \STATE $\mathbf{P_{origin}} = CROP(\mathbf{I}, \text{Coord}(\mathbf{P^{2D}_i)})$;
    \STATE $\mathbf{P_{new}} =  \alpha \mathbf{P_{origin}} + (1 - \alpha) \mathbf{P^{2D}_{i}}$;
    \STATE $\mathbf{I} \leftarrow PASTE(\mathbf{I}, \mathbf{P_{new}})$
\ENDFOR
\ENSURE $\mathbf{P}, \mathbf{I}$
\end{algorithmic}
\label{algo:gt_aug}
\end{algorithm}
%

Depth-Aware GT-AUG simply follows the augmentation strategy in the 3D domain, but at the same time, keeps the synchronization in the image plane through MixUp-based cut-and-paste. The key intuition is that the MixUp technique does not fully remove the corresponding information after pasting augmented patches on top of the original 2D image. On the contrary, it decays the compactness of such information with respect to the depth to guarantee the existence of the feature from the corresponding points. Concretely, if one object is occluded by other instances $n$ times, the transparency of this object region will be decayed by a factor of $ (1 - \alpha)^n$ according to its depth order. 
 
\subsection{Image-Level Dropout Training Strategy}

\begin{figure}
\subfloat[Vanilla Image Fusion]{
  \begin{minipage}[t]{0.5\linewidth}
    \centering
    \includegraphics[scale=0.45]{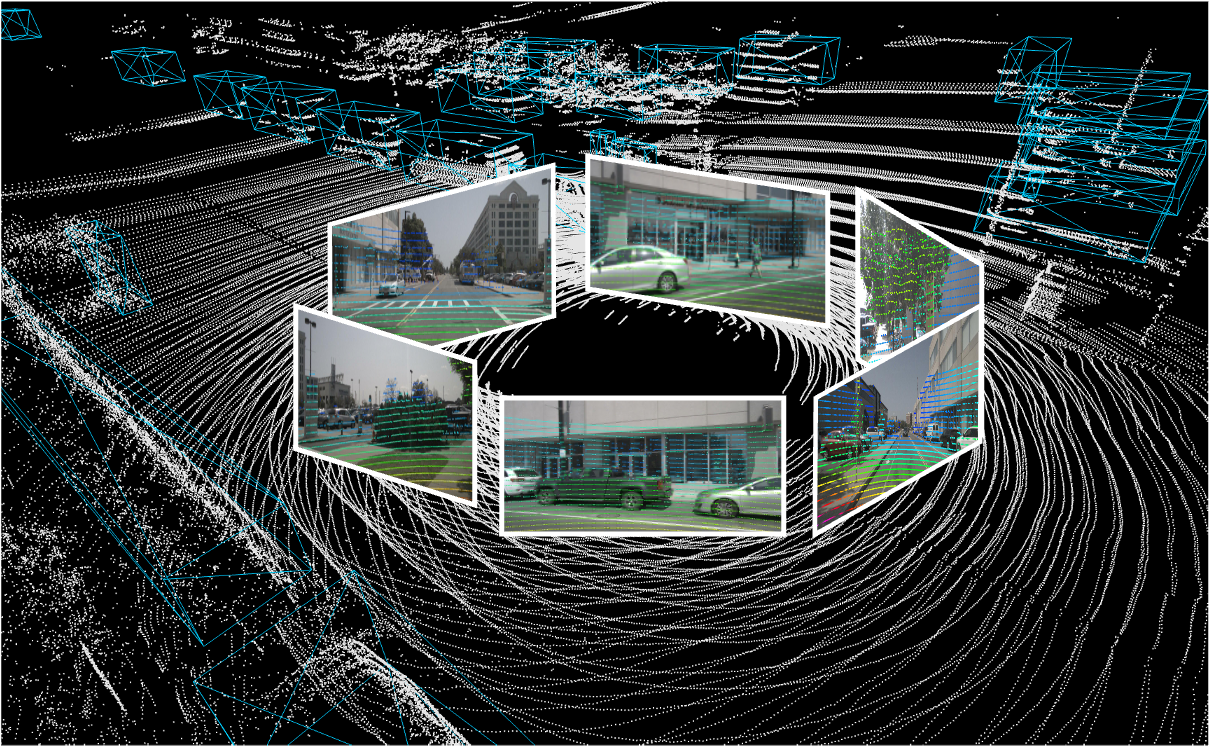}
    \label{fig:side:a}
  \end{minipage}%
}
\subfloat[Image-Level Dropout Fusion]{
  \begin{minipage}[t]{0.45\linewidth}
    \centering
    \includegraphics[scale=0.45]{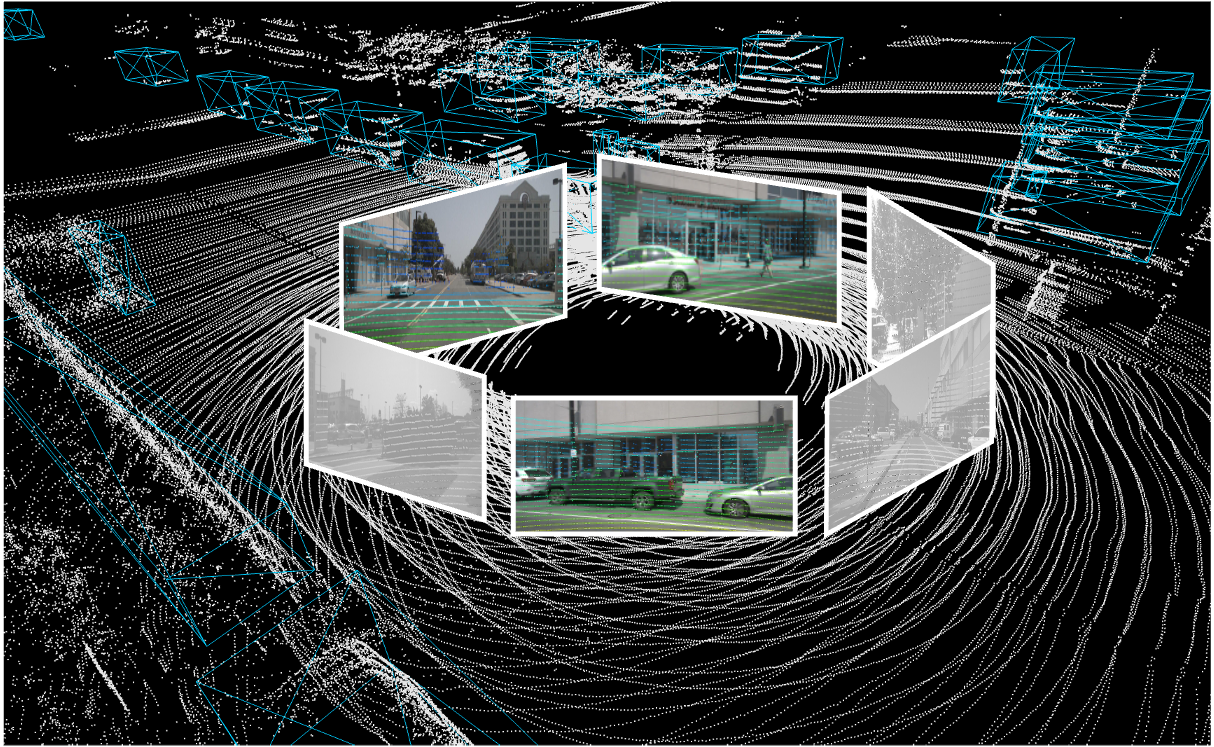}
    \label{fig:side:b}
  \end{minipage}
}
\caption{Visualization of our proposed image-level dropout training strategy compared to the vanilla fusion method. We enable the model to acquire ad-hoc inference by randomly blinding several cameras during training. The images in while-black (b) denote the dropout RGB images where we pad them with zeros for fusion.}
\label{fig:dropout}
\end{figure}

Actually, image is usually an optional input and may not be supported in all 3D detection systems. Therefore, a more realistic and applicable solution to multi-modal detection should be in a dynamic fusion manner: when images are unavailable, the model detects objects based on raw point clouds; when images are available, the model conducts feature fusion and yields better prediction. To achieve this goal, we propose an image-level dropout training strategy by randomly dropping the aggregated image features at the image level and padding them with zeros during training, as shown in Figure~\ref{fig:dropout}. Since the imagery information is intermittently missed, the model should gradually learn to utilize 2D features as one alternative input. Later, we will show that such a strategy not only speeds up the training speed greatly (with fewer images to process per batch) but also improves the final performance. 

\section{Experiments}

\subsection{Dataset and Experimental Setup}

\textbf{Dataset.}
The nuScenes dataset \cite{caesar2020nuscenes} is one of the most popular datasets for 3D object detection, consisting of 700 scenes for training, 150 scenes for validation, and 150 scenes for testing. For each scene, it includes 6 camera images to cover the whole viewpoint. In terms of the overlapping regions between images, we predefine the image fetching priority sequence to avoid the ambiguous problem.  \\
\textbf{Experimental Setup.}
We select Object DGCNN~\cite{wang2021object} and CenterPoint~\cite{yin2021center} as 3D base detectors for the nuScenes dataset. For the image branch, we adopt a light-weight backbone CSPNet~\cite{wang2020cspnet}, the same one used in YOLOX-Tiny~\cite{ge2021yolox}, as the feature extractor, followed by PAFPN~\cite{liu2018path}. We also pretrain the image branch with 2D detection supervision on nuImages by adding an extra head~\cite{ge2021yolox}. The voxel size is set to $(0.1m, 0.1m, 0.1m)$ if not specified. To avoid the redundant computational cost, we adopt dynamic voxelization~\cite{zhou2020end} to reduce the number of voxel features. As for the DeformCAFA module, the head number is set to 4 and the deformable point is set to 8. All the feature pyramid layers share the same weight for the feature aggregation operation. All runtimes are measure on a NVIDIA V100 GPU.
The whole framework is optimized with hybrid optimizers in an end-to-end manner. The 3D branch is optimized with AdamW and the 2D branch is optimized with SGD. We use MMDetection3D \cite{mmdet3d2020} as our codebase and apply the default settings, if not specified. 

\subsection{Main Results}

\subsubsection{Results on 3D Object Detectors}

We first implement AutoAlignV2 on two representative 3D detector baselines: CenterPoint (anchor/center-based) and Object DGCNN (transformer-based) on nuScenes validation subset. The final performance is reported in Table \ref{tab:nus_val}. Our AutoAlignV2 greatly boosts its vanilla 3D baselines by 3.7/4.5 on mAP and 2.4/2.4 on NDS score, respectively. This validates the effectiveness and generalization of the proposed method under different 3D detection frameworks.

\begin{table}[!h]
	\centering
	\caption{Comparison of detection results based on Object DGCNN and CenterPoint with and without AutoAlignV2 on nuScenes validation subset.}
	\label{tab:nus_val}
	\setlength{\tabcolsep}{2.0mm}
		\begin{tabular}{c |c | c c }
			\toprule
			Method & AutoAlignV2 & mAP & NDS \\
			\midrule
			\midrule
			Object DGCNN\cite{wang2021object} & & 60.73 & 67.14  \\
			Object DGCNN\cite{wang2021object} & $\checkmark$ & \textbf{64.42} & \textbf{69.52}\\
			\midrule
			CenterPoint \cite{yin2021center} & & 62.56 & 68.84 \\
			CenterPoint \cite{yin2021center} & $\checkmark$ & \textbf{67.05} & \textbf{71.23}  \\
			\bottomrule
		\end{tabular}
\end{table} 

\subsubsection{Comparison with State-of-the-Arts}
In addition to offline results, we also report the detection performance on nuScenes test leaderboard compared to various detection approaches. The results are shown in Table \ref{tab:nus_test}. Our final model is based on CenterPoint with a voxel size of $(0.075m, 0.075m, 0.2m)$. It surpasses all the other counterparts including the recently developed MoCa~\cite{zhang2020exploring} and PointAugmenting~\cite{wang2021pointaugmenting} by roughly 2.0 mAP, achieving new state-of-the-arts on this competitive benchmark. When observing the results in detail, we can find that the construction vehicle, motorcycle, and bicycle are separately improved by 13.1, 13.4, and 17.4 mAP. Such huge enhancements manifest the superiority of our proposed AutoAlignV2 to deal with hard-to-detect examples. 
\begin{table*}[!t]
	\label{tab:mvx_nus_val}
    \begin{center}
    \caption{Comparison with previous methods on nuScenes test leaderboard. ``C.V.'' and ``Ped.'' are the abbreviations of construction vehicle and pedestrian, respectively. NDS score, mAP, and APs of each category are reported. The single class AP not reported in the paper is marked by ``-''. The best results are highlighted in bold.
	}
	\label{tab:nus_test}
		\begin{tabular}{c|cc|cccccccc}
		    \bottomrule
            Method & NDS &mAP &Car& Truck & Bus & Trailer & C.V. & Ped. & Motor & Bicycle  \\ 
            \hline
            \hline
            3D-CVF~\cite{yoo20203d} &49.8&42.2&79.7&37.9&55.0&36.3&-&71.3&37.2&- \\
            PointPainting~\cite{vora2020pointpainting} &58.1&46.4&77.9&35.8&36.1&37.3&15.8&73.3 & 41.5 & 24.1\\
            CVCNet~\cite{chen2020every} & 66.6 & 58.2 & 82.6 & 49.5& 59.4 & 51.1 & 16.2 & 83.0 & 61.8 & 38.8\\
            AFDetV2~\cite{zhu2019class} & 68.5 & 62.4 & 86.3 & 54.2 & 62.5 & 58.9 & 26.7 & 85.8 & 63.8 & 34.3\\
            MVP~\cite{yin2021center} & 70.5 & 66.4 & 86.8 & 58.5 & 67.4 & 57.3 & 26.1 & 89.1 & 70.0 & 49.3 \\
            MoCa~\cite{zhang2020exploring} &70.9 & 66.6 & 86.7 & 58.6 & 67.2 & 60.3 & 32.6 & 87.1 & 67.8 & 52.0  \\
            AutoAlign~\cite{chen2022autoalign} & 70.9 & 65.8 & 85.9 & 55.3 & 67.7 & 55.6 & 29.6 & 86.4 & 71.5 & 51.5 \\
            PointAugmenting~\cite{wang2021pointaugmenting} & 71.1 & 66.8 & 87.5 & 57.3 & 65.2 & 60.7 & 28.0 & 87.9 & 74.3 & 50.9 \\
            \hline
            CenterPoint~\cite{yin2021center} & 67.3 & 60.3 & 85.2 & 53.5 & 63.6 & 56.0 & 20.0 & 84.6 & 59.5 & 30.7 \\
            AutoAlignV2 (Ours) & \textbf{72.4} & \textbf{68.4} & 87.0 & 59.0 & 69.3 & 59.3 &33.1 & 87.6 & 72.9 & 52.1\\
            \bottomrule
		\end{tabular}
	\end{center}
\end{table*}

%

\subsection{Ablation Studies}

In this section, we provide extensive ablations to gain a deeper understanding of AutoAlignV2. For efficiency, 1/8 nuScenes training set is used. 

\subsubsection{Ablation Studies on AutoAlignV2}

To understand how each component in AutoAlignV2 facilitates the detection performance, we test each module independently on the baseline detector: CenterPoint and report its performance in Table \ref{tab:ablations}. The overall mAP score starts from 50.3. When we add the Cross-Domain DeformCAFA module together with the image branch, the mAP score is raised by 6.7\%. Such a significant improvement validates the correctness of the incorporation of image features and the effectiveness of the proposed deformable feature alignment module. Then, we adopt the image-level dropout strategy to improve the training speed. The performance does not drop and is even slightly improved by another 0.1 mAP. When the depth-aware GT-AUG is added, the accuracy is further promoted by 1.4 mAP. Although the improvement is not remarkable, depth-aware GT-AUG greatly simplifies the synchronization process in the joint image-point augmentation. 
\begin{table*}[!hbpt]
	\centering
		\caption{Effect of each component in our AutoAlignV2. Results are reported on nuScenes validation set with CenterPoint.}
		\begin{threeparttable}
		\setlength{\tabcolsep}{1.0mm}
			\begin{tabular}{c c c | c c }
				\toprule
				DeformCAFA~~~ & Image-level Dropout~~~ & Depth-aware GT-AUG~ & mAP & NDS \\
				\midrule
			    \midrule
				    & & & 50.28 & 58.71\\
				    \checkmark & & & 56.96 & 62.54\\
				    \checkmark & \checkmark & & 57.03 & 62.52\\
				    \checkmark & \checkmark & \checkmark & \textbf{58.45} & \textbf{63.16}\\
				\bottomrule
			\end{tabular}
		\end{threeparttable}
	\label{tab:ablations}
\end{table*}

\subsubsection{Ablation Studies on Cross-Domain DeformCAFA}

\textbf{Comparison with other fusion mechanisms.} In this experiment, we keep all settings the same except for the cross-modal feature fusion method for a fair comparison. We consider the following strategies used in PointPainting \cite{sindagi2019mvx}, MoCa \cite{zhang2020exploring}, AutoAlign \cite{chen2022autoalign}, and PointAugmenting \cite{wang2021pointaugmenting}, and compare them with Cross-Domain DeformCAFA in Table \ref{tab:abla_fusion}. It can be found that AutoAlignV2 outperforms all the other fusion mechanisms by a large margin, verifying the effectiveness of our proposed approach. The enhancement mainly stems from two aspects: (i) multi-level features are fully utilized thanks to the optimization of computational complexity and (ii) superiority of relational modeling on cross-domain features across different modalities.
	\begin{table}[!t]
 \centering
 \caption{Comparison with different feature fusion strategies adopted in current multi-modal detectors. Methods with * indicate our own implementation.}
  \begin{threeparttable}
  \setlength{\tabcolsep}{2.0mm}
   \begin{tabular}{c | cc }
    \toprule
    Fusion Strategy & mAP & NDS \\
    \midrule
       \midrule
       Baseline w/o Img & 50.28 & 58.71 \\
    PointPainting*~\cite{sindagi2019mvx} & 55.45 & 61.44\\
    MoCa~\cite{zhang2020exploring} & 55.91 & 61.54\\
    AutoAlign~\cite{chen2022autoalign} & 56.69 & 61.93\\
    PointAugmenting*~\cite{wang2021pointaugmenting} & 56.75 & 62.11\\
    Cross-Domain DeformCAFA & \textbf{58.45} & \textbf{63.16}\\
    \bottomrule
   \end{tabular}
  \end{threeparttable}
 \label{tab:abla_fusion}
\end{table}

\noindent\textbf{Strategies on token generation.} To validate the necessity of the cross-domain token generation, we compare our method with various policies: generated from voxel features only, image features only, and their combinations including concatenation, addition, and multiplication. As given in Table \ref{tab:abla_token}, utilizing the voxel features as the query token cannot guarantee satisfying results, since 3D features can hardly perceive information in the interaction between cross-modal features. The result produced by the image features is also limited, possibly due to the lack of information from 3D points. The performance of simply concatenating or adding them together remains poor. We infer the reason that though both features contain the same identity information, it is still hard for the model to figure them out when blending with the domain-specific representation. Finally, we obtain the best performance with the multiplication version, which proves the assumption in Section \ref{sec:cross_token}.
	\begin{table}[!h]
	\centering
	\caption{Ablations on different strategies in query token generation for Cross-Domain DeformCAFA module. ``Operation'' denotes the interact operation between the points and image features to generate tokens.}
	\label{tab:abla_token}
		\begin{threeparttable}
			\begin{tabular}{c c c | c c}
				\toprule
				Pts Feat & Img Feat & Operation & mAP & NDS \\
				\midrule
			    \midrule
			    \checkmark & & - & 57.10 & 61.77\\
			     & \checkmark & - & 57.77 & 62.08\\
			    \midrule
			    \checkmark & \checkmark & Concat & 58.01 & 62.32\\
			    \checkmark & \checkmark & Add & 57.94 & 62.13\\
			    \checkmark & \checkmark & Multiply & \textbf{58.45} & \textbf{63.16}\\
				\bottomrule
			\end{tabular}
		\end{threeparttable}
\end{table}


\subsubsection{Ablation Studies on Depth-Aware GT-AUG}

\textbf{Comparison with other cross-modal GT-AUG.} We compare depth-aware GT-AUG together with other cross-modal data augmentation approaches proposed in MoCa~\cite{zhang2020exploring} and PointAugmenting~\cite{wang2021pointaugmenting}. As shown in Table \ref{tab:abla_aug}, the depth-aware GT-AUG slightly surpasses all the other strategies even without point filtering or 2D occlusion checking, which greatly overcomes the difficulty in cross-domain synchronization. Moreover, we can see from Figure ~\ref{fig:aug_img} that the depth-aware GT-AUG is able to produce smoother images for image fusion, which enhances the quality of 2D features during the cross-modal fusion process.
		
\begin{table}[!h]
 \centering
 \caption{Comparison with various cross-modal GT-AUG strategies. ``Occ Check'': abandoning the instance paste if it has certain overlap with the original instances in the images; ``Pts Filter'': filtering the points to guarantee that points of one instance will not aggregate the image features from another occluded one.}
  \begin{threeparttable}
  \setlength{\tabcolsep}{0.7mm}
   \begin{tabular}{c |c c| cc }
    \toprule
    Method & Occ Check & Pts Filter & mAP & NDS \\
    \midrule
    \midrule
    	w/o Aug & & & 40.12 & 45.39\\
       MoCa~\cite{zhang2020exploring} &\checkmark &  & 53.08 & 56.54\\
       Wang et.al~\cite{wang2021pointaugmenting}~~ & & \checkmark & 53.16 & 56.91\\
       DA-GTAUG & & & \textbf{53.48} & \textbf{57.16}\\
    \bottomrule
   \end{tabular}
  \end{threeparttable}
 \label{tab:abla_aug}
\end{table}

\begin{figure}[ht!]
\centering
   \includegraphics[width=0.5\textwidth]{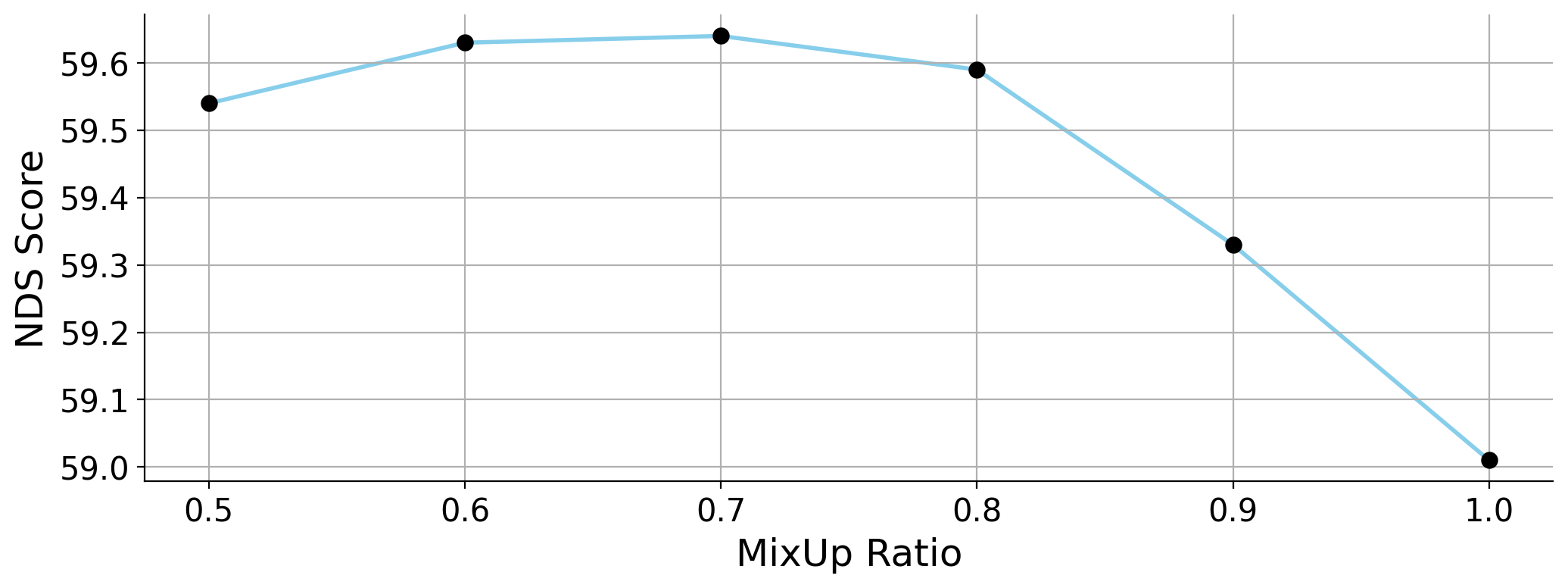}
   \vskip -1em
\caption{Ablation study on the mix-up ratio $\alpha$ introduced in depth-aware GT-AUG.}
\label{fig:mixup}
\end{figure}

\noindent\textbf{GT-AUG Mix-up Ratio.} In Figure \ref{fig:mixup}, we investigate how the mix-up ratio $\alpha$ in the depth-aware GT-AUG affects the model performance. It can be seen that the detection result is not sensitive to the mix-up ratio ranging from 0.5 to 0.8, where the NDS only fluctuates within 0.1\%. However, the score drops about 0.7 mAP with $\alpha=1.0$, where the depth-aware GT-AUG degenerates to the original GT-AUG implementation in MoCa \cite{zhang2020exploring}. Since no occlusion checking or point filtering is performed, points may be fused with other imagery information, leading to the ambiguous learning issue.

\subsubsection{Ablation Studies on Image-level Dropout.}

Considering that AutoAlignV2 can be dynamically trained with or without images, namely dynamic image fusion, we study such an attribute and how it contributes to the final performance. Concretely, we vary the number of images for training in our image-level dropout strategy and report the detection accuracy as well as the training time in Table \ref{tab:abla_adhoc}. From the table, we can find that reducing the number of training images from 6 to 3 has little effect on the performance of the model but greatly reduces the training time by 1.5$\times$. However, if continuously reducing this number to 1, the performance incurs an evident decline. We infer the reason that single image training is not enough for fully cross-modal fusion learning. Therefore, we adopt 3 images per scene in our experiments.
	
\begin{table}[!h]
 \centering
 \caption{Ablation studies on the number of images for fusion during the training process with our proposed image-level dropout strategy.}
  \begin{threeparttable}
   \begin{tabular}{c c | cc }
    \toprule
    ~~~\# Images~~ & ~~Training Time~~~ & ~~~mAP~~~ & NDS~~~ \\
    \midrule
       \midrule
       0 & 7.6h & 50.28 & 58.71\\
       1 & 8.5h & 57.93 & 62.84\\
       3 & 9.7h & 58.45 & \textbf{63.16}\\
       6 & 14.1h & \textbf{58.51} & 63.11\\
    \bottomrule
   \end{tabular}
  \end{threeparttable}
 \label{tab:abla_adhoc}
\end{table}

\subsection{Dynamic Inference and Runtime}

Autonomous driving is a direct application of multi-modal 3D object detection. Therefore, the practicality and inclusiveness of the model are also vital. As mentioned in Section~\ref{sec:cross_token}, AutoAlignV2 fits to different inference modes, no matter the images are available or not. We carefully measure the inference performance of AutoAlignV2 under different settings and report its runtime per frame in Table \ref{tab:runtime}. Compared with the LiDAR-only detector: CenterPoint, our AutoAlignV2 takes only 123 ms for the extra 2D image branch, thanks to its light-weight backbone: CSPNet. We resize all the images to 640$\times$1280 for efficient fusion.  In addition to fully surrounding images for cross-modal fusion, our method is also qualified for the LiDAR-only scenarios without any extra computational cost compared to vanilla CenterPoint, but still maintains the detection accuracy.

\begin{table}[!h]
 \centering
 \caption{Inference time of AutoAlignV2 on nuScenes dataset. ``\# Images'' means the number of images to load during inference. }
 \setlength{\tabcolsep}{1.0mm}
  \begin{threeparttable}
   \begin{tabular}{c | c c cc }
    \toprule
    Method & \# Images & Inference Time & mAP & NDS \\
    \midrule
    \midrule
    CenterPoint & - & 85ms & 50.28 & 58.71\\
    AutoAlignV2 & 6 & 208ms & \textbf{58.45} & \textbf{63.16} \\
    AutoAlignV2 & 3 & 181ms & 54.32 & 60.84\\
    AutoAlignV2 & 0 & 87ms & 50.29 & 58.67 \\
    \bottomrule
   \end{tabular}
  \end{threeparttable}
 \label{tab:runtime}
\end{table}

\section{Conclusion}

In this paper, we develop a dynamic and fast multi-modal 3D object detection framework, AutoAlignV2. It greatly speeds up the fusion process by utilizing multi-layer deformable cross-attention networks to extract and aggregate features from different modalities. We also design the depth-aware GT-AUG strategy to simplify the synchronization between 2D and 3D domains during the multi-modal data augmentation process. Interestingly, our AutoAlignV2 is much more flexible and can infer with and without images in an ad-hoc manner, which is more suitable for the real-world systems. 
We hope AutoAlignV2 can serve as a simple yet strong paradigm in multi-modal 3D object detection.

\noindent{ \bf Acknowledgments:} This work was supported by the USTC-NIO Joint Research Funds KD2111180313. We acknowledge the support of GPU cluster built by MCC Lab of Information Science and Technology Institution, USTC.

{\small
\bibliographystyle{ieee_fullname}
\bibliography{aav2.bib}
}

\end{document}